\documentclass{article}

\usepackage[margin=1in]{geometry} 

\usepackage[english]{babel}
\usepackage{amsthm}
\usepackage{titling}
\usepackage{amssymb}
\usepackage{mdframed}
\usepackage[utf8]{inputenc}
\usepackage{algorithm}
\usepackage{algpseudocode}
\usepackage{graphicx} 
\usepackage{booktabs} 
\usepackage{amsmath} 
\usepackage{authblk}
\usepackage{subcaption} 
\newcounter{subsubsubsection}[subsubsection]

\newmdtheoremenv[linewidth=1.5pt,
                 topline=true,
                 bottomline=true,
                 rightline=false,
                 leftline=false,
                 innertopmargin=\baselineskip,
                 innerbottommargin=\baselineskip,
                 skipabove={\dimexpr0.5\baselineskip+\topsep\relax},
                 skipbelow={\dimexpr0.5\baselineskip+\topsep\relax},
                 innerleftmargin=10pt,
                 innerrightmargin=10pt,
                 splittopskip=\topskip,
                 splitbottomskip=0.3\topskip,
                 frametitleaboveskip=1\baselineskip,
                 frametitlebelowskip=1\baselineskip]{thmd}{Theorem}

\title{Iterative Causal Segmentation\thanks{Disclaimer: The opinions expressed in this presentation are those of the presenters and do not necessarily reflect the views of AstraZeneca. The analyses presented in this presentation are based on Uber’s open-source CausalML GitHub repository data and do not represent AstraZeneca data.}\\
 \small Filling the Gap between Market Segmentation and Marketing Strategy}

\author[1]{Kaihua Ding}
\author[1]{Jingsong Cui}
\author[1]{Mohammad Soltani}
\author[1]{Jing Jin}

\affil[1]{AZ Brain, AstraZeneca PLC}

\date{}
\begin{document}

\renewcommand{\thefootnote}{\texttt{+}} 

\maketitle

\begin{abstract}
The field of causal Machine Learning (ML) has made significant strides in recent years. Notable breakthroughs include methods such as meta learners~\cite{Künzel2019} and heterogeneous doubly robust estimators~\cite{kennedy2023optimal} introduced in the last five years. Despite these advancements, the field still faces challenges, particularly in managing tightly coupled systems where both the causal treatment variable and a confounding covariate must serve as key decision-making indicators. This scenario is common in applications of causal ML for marketing, such as marketing segmentation and incremental marketing uplift. In this work, we present our formally proven algorithm, iterative causal segmentation, to address this issue.

\end{abstract}

\section{Motivation}

The integration of machine learning into market segmentation has significantly transformed the development of marketing messages and strategies. However, categorizing individuals into rigid market segments such as 'loyalists' or 'dabblers' fails to account for the dynamic nature of consumer circumstances, potentially leading to ineffective marketing and wasted resources. This underscores the limitations of relying solely on traditional market segmentation for marketing actions, a challenge initially highlighted by Wendell R. Smith in the 1950s~\cite{smith1956marketsegmentation}. Despite its innovative approach, market segmentation's static methodology struggles to capture evolving consumer behaviors, risking oversimplification and ineffective strategy development.

The emergence of causal inference in marketing provides a solution by identifying the causative factors behind consumer behaviors, enabling the development of predictive and effective marketing strategies with methods like Uplift Trees and Meta Learners~\cite{Radcliffe2011Uplift, Künzel2019, Zhao2017Uplift}. However, these approaches often overlook the complexity that arises when market segmentation becomes an intertwined part of the promotional market, acting both as a significant confounder and a desired output for marketing purposes, especially in contexts like pharmaceutical call planning and broadcasting media. Causality analysis alone cannot address it as both a confounder and an output.

To tackle these challenges, we propose the iterative causal segmentation algorithm, which merges causal inference with market segmentation to surmount their individual limitations. This method not only provides a nuanced understanding of consumer behaviors but also amplifies the effectiveness of marketing efforts across various segments. Nonetheless, in marketing scenarios where the interdependency between segmentation and causality analysis poses a distinct challenge, leading to a cyclical problem where each influences the outcome of the other, our proposed solution aims to harmonize these methodologies. By leveraging their strengths, we seek to refine marketing strategies effectively.

\section{Iterative Causal Segmentation}
\subsection{Average Treatment Effect (ATE)}

A key challenge for marketing professionals concerns generating a sales uplift from promotional campaigns, quantified by:
\begin{equation}
    \text{Expected uplift gain from applying promotion} = E(Y^{A=1}) - E(Y^{A=0}),
    \label{eqn:ate}
\end{equation}

where:

\begin{itemize}
    \item \(A\) represents the treatment, i.e., the promotional campaign.
    \item \(Y\) is the outcome, indicating the effect of the campaign.
\end{itemize}

The difference between these two expected values represents the expected uplift per person due to the promotion. Equation~\ref{eqn:ate} represents the expected average purchase uplift if the promotion is applied. This expected uplift is also known as the Average Treatment Effect (ATE) in causality analysis. The Average Treatment Effect is defined as:

\begin{equation}
\text{ATE} = E(Y^{A=1}) - E(Y^{A=0})
\label{eqn:ate_math}
\end{equation}

This formulation allows us to quantitatively assess the overall impact of the treatment across the entire population. Additionally, recently developed causal machine learning techniques are capable of attributing uplift gain to individual samples \cite{hansotia_2002, Künzel2019} by estimating the conditional average treatment effect.

\subsection{Conditional Average Treatment Effect (CATE)}

The Conditional Average Treatment Effect, which can also be understood as the expected individual treatment effect (ITE) conditional on covariates \(X\), is defined as:

\begin{equation}
\text{CATE} = E(Y^{A=1} | X) - E(Y^{A=0} | X)
\label{eqn:cate}    
\end{equation}

where:

\begin{itemize}
    \item \(A\) represents the treatment, i.e., the promotional campaign.
    \item \(Y\) is the outcome, indicating the effect of the campaign.
    \item \(X\) represents the covariates or features of the individual units that might affect how the treatment impacts the outcome.
\end{itemize}

CATE provides a more nuanced understanding of how the effect of the treatment varies across different segments of the population based on the covariates \(X\). It allows for the estimation of how the treatment effect differs among individuals or specific groups within the population, enabling targeted interventions.

\subsection{Causal Graph}

A causal graph, often used in the context of causal inference and statistics, is a graphical representation that models the causal relationships between variables. It is a type of directed graph where nodes represent variables (e.g., events, conditions, or quantities), and edges represent causal effects from one variable to another. This concept is foundational in understanding causal inference, allowing researchers to visually represent and analyze the cause-and-effect relationships within a system.

We can express the relationship among promotion application \(A\), marketing segmentation \(S\), covariates \(X_i\), and promotion outcome \(Y\) as the following causal graph.

\begin{figure}[h!]
\centering
\includegraphics[width=0.5\linewidth]{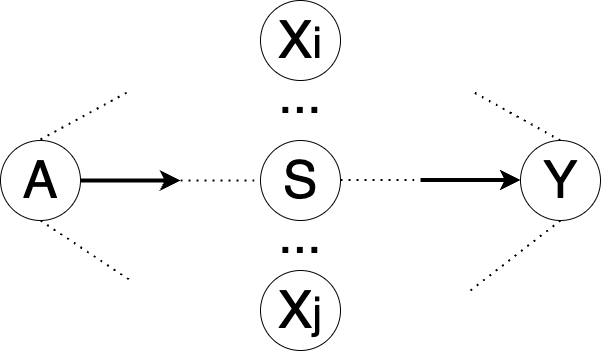}
\caption{Potential causal graph, where promotion is \(A\), covariates are \(X_i\), promotion outcome is \(Y\), and marketing segments \(S\) is a confounder.}
\label{fig:causal_graph}
\end{figure}

The causal graph provides an illustrative relationship among treatment \(A\), covariates \(X\), and outcome variables \(Y\)~\cite{pearl2009causality}. A causal graph is a product of a hypothesis and rationalization, rather than ground truth. Regardless, the causal graph drawn according to empirical knowledge needs to be verified through sensitivity analysis. Rather than focusing on how to come up with the best causal graph, the goal of drafting a causal graph should be to guide the collection of covariates and the necessary control variables for covariate matching. In the authors' opinion, establishing a causal graph is helpful but not the most critical aspect of causality analysis. The causal graph should change and needs to change, especially if the causal relationship contains human perspective or ambiguities. It's the overall reliability in the sensitivity analysis that truly matters as the end result.

In addition, the main usage of a causal graph is to assist users in how to control for confounders or covariates to ensure that the causality analysis is valid, e.g., backdoor and frontdoor path criteria~\cite{Pearl2012DoCalculus}. For simplicity, we use the disjunctive criterion~\cite{VanderWeele2019ConfounderSelection} in this paper and we perform sensitivity analysis to check the reliability of our causal analysis, which also serves as an assessment for the impact of unobserved confounders and uncertainty quantification.

\subsection{The Iterative Causal Segmentation Algorithm}

\begin{figure}[h!]
\centering
\includegraphics[width=\linewidth]{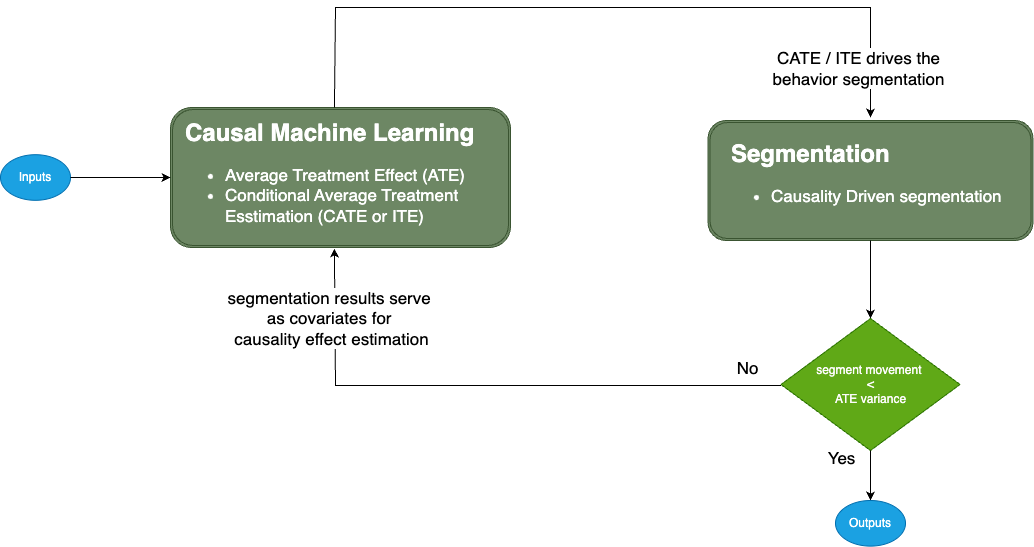}
\caption{The diagram description on how to use causal machine learning to guide segmentation efforts. The goal is to achieve the joint convergence of the causal machine learning module as well as the convergence of the segmentation module.}
\label{fig:workframe}
\end{figure}

In this section, we illustrate our proposed algorithm, Iterative Causal Segmentation. The key challenge to address is the tightly coupled nature between segmentation and causality analysis. Instead of shying away from the tightly coupled nature of these two computational modules, we propose a joint convergence method that solves for segmentation and causality analysis simultaneously, as shown in Figure~\ref{fig:workframe}.

Figure~\ref{fig:workframe} figuratively describes the joint convergence algorithm's workflow, explicitly considering the tightly coupled nature and mutual influence of the two modules. For the purpose of effective uplift behavior segmentation, the causal machine learning module will generate useful uplift incremental estimation, which we will feed into the segmentation module to produce segmentation that closely reflects the promotion uplift effect. Segmentation results will then serve as input in the causal machine learning, and the overall system is considered converged if the amount of segment movement becomes less than the population size variance determined by ATE variance estimates. This workflow can be more formally defined as follows in Algorithm~\ref{alg:causal_segmentation}.

\begin{algorithm}[h!]
\caption{Causal Machine Learning Driven Segmentation}\label{alg:causal_segmentation}
\begin{algorithmic}[1]
\Require{Matched treatment / control variables \( A \), outcome \( Y \), initial segmentation \( S \), and covariates \( X \)}
\Ensure{Causal Data Assumptions}
\Statex
\State \(\textrm{Segment Movement} \gets \textrm{initial value}\)
\State \(\textrm{ATE Variance} \gets \textrm{initial value}\)
\While{\(\textrm{Segment Movement} > \left( \textrm{ATE Variance} \times \textrm{Population Size}\right)\)} 
    \State \( \textrm{ATE} \gets \Call{ComputeATE}{A, S, Y, X} \)
    \State \( \textrm{ATE Variance}  \gets \Call{EvaluateATEVariance}{\textrm{ATE}} \)
    \State \( \textrm{CATE} \gets \Call{ComputeCATE}{A, S, Y, X} \)
    \State \( S \gets \Call{Segmentation}{\textrm{CATE}} \)
    \State \( \textrm{Segment Movement} \gets \Call{EvaluateSegmentMovement}{S} \)
\EndWhile
\State \textbf{return} \( S \)
\end{algorithmic}
\end{algorithm}

Algorithm~\ref{alg:causal_segmentation} formally describes Figure~\ref{fig:workframe} in a more concise and concrete pseudocode fashion. Observing Algorithm~\ref{alg:causal_segmentation}, one might question whether there is any need to go to such great lengths to formulate a new algorithm to solve the causality behavior segmentation problem in marketing. This is also the question that the authors are interested in solving first. Are there alternatives to the simpler segmentation algorithm that serve the same purpose without the need to solve the coupled segmentation and causality system? We can formally prove that such an alternative segmentation method does not exist. The causality uplift segmentation analysis is exclusively determined by the iterative causal segmentation algorithms.

We formalize the proof as the following statement of Causal Segmentation Exclusivity. If this theory holds true, it means our iterative causal segmentation is necessary since segmentation obtained this way is exclusive to the causality.

\begin{thmd}[Causal Segmentation Exclusivity]
\label{thmd:exclusivity}
Let us define the Conditional Average Treatment Effect (CATE), also known as the Individual Treatment Effect (ITE), as follows:
\begin{equation}
    \textrm{CATE}(X) = \mathbb{E}[Y^{A=1} - Y^{A=0} \ |\ X]
\end{equation}
where \( Y^{A=1} \) represents the potential outcome if treated or received campaign promotion, \( Y^{A=0} \) represents the potential outcome if not treated or not receiving campaign promotion, and \( X \) is a vector of individual characteristics.

Segmentation, in the context of causal machine learning, is the process of partitioning a population into distinct groups based on their respective CATE estimates. Here, the segmentation function \( S \) is exclusively dependent on \( CATE(X) \):
\begin{equation}
    S = f(\textrm{CATE}(X))
\end{equation}

We hypothesize that no other factors other than CATE influence the segmentation, that is, segmentation is a function of CATE alone.
\end{thmd}

A formal proof through contradiction is detailed below, 

\renewcommand\qedsymbol{$\blacksquare$}
\vspace{3pt}
\begin{proof}[\textbf{Proof of Theorem 1}]
Assume for contradiction that there exists another driver \( D \) which influences the segmentation such that:
\begin{equation}
    S' = f(\textrm{CATE}(X), D)
\end{equation}
where \( S' \) represents a new segmentation outcome due to the presence of driver \( D \).

If driver \( D \) were to affect the segmentation, we would observe a change in the segmentation outcome \( S \) without a corresponding change in the CATE estimates. This would contravene the initial assumption that segmentation is exclusively driven by CATE.

Since our operational framework stipulates CATE as the sole driver for segmentation, the supposition of another influencing driver \( D \) is invalid. Therefore, changes in segmentation are directly correlated with changes in the CATE estimates, thus affirming that the segmentation is indeed a causal behavior segmentation when CATE is the exclusive driver.
\label{pf:exclusitivity}
\end{proof}

The Proof~\ref{pf:exclusitivity} formally proves that the iterative causal segmentation is a necessary system to address when the tightly coupled nature between causality and segmentation exists. The significance of this proof lies in its affirmation for the development of such algorithms. Now that we have justified the legitimacy of developing this new algorithm, Algorithm~\ref{alg:causal_segmentation}, we will evaluate its performance in the next section.


\section{Results and Discussion}
\subsection{Data Sources Disclaimer and Discussion}
We examine the performance of Algorithm~\ref{alg:causal_segmentation} by applying it to open-source data from the Uber CausalML package~\cite{chen2020causalml}. We want to emphasize that all causal machine learning algorithms derive their origin from causality analysis. As a result, all the data assumptions required for performing causality analysis need to be true to ensure the comprehensiveness and validity of the analysis, as outlined in Table~\ref{table:assumptions}.

\begin{table}[h!]
\centering
\caption{Causal Data Assumptions. In the table below, subscript indices denote sample indices, and superscript indicates the treatment/control variable.}
\label{table:assumptions}
\begin{tabular}{@{}lp{8cm}@{}}
\toprule
\textbf{Assumption} & \textbf{Expression} \\ \midrule
SUTVA & Each unit sample \(i\)'s potential outcomes \(Y_i^A\) are unaffected by the treatment assignment \(A_j\) of any other unit \(j\), where \(A=1\) indicates treatment is assigned and \(A=0\) indicates control. \\
Ignorability & \(Y_i^{A=0}, Y_i^{A=1} \perp A_i | X_i\) for all units \(i\), where \(A_i\) is the treatment assignment and \(X_i\) are covariates. \\
Positivity & \(P(A_i = a | X_i = x) > 0\) for all levels of treatment \(a\) and covariates \(x\). \\
Consistency & If \(A_i = a\), then the observed outcome \(Y_i\) is equal to the potential outcome \(Y_i^{A=a}\). \\ \bottomrule
\end{tabular}
\end{table}

The SUTVA assumption concerns the principle that the treatment received by one unit does not affect the outcomes of any other unit. In other words, the potential outcome for any individual is assumed to be independent of the treatment assignments of all other individuals. The ignorability assumption, also known as the conditional independence assumption or no unmeasured confounders assumption, plays a pivotal role in the field of causal inference, particularly in observational studies where random assignment to treatment and control groups is not feasible. This assumption is crucial for estimating causal effects from observational data, where the potential for confounding variables is a significant concern. The ignorability assumption allows researchers to control for confounding variables through statistical methods such as regression, matching, stratification, or weighting. By adjusting for a comprehensive set of observed covariates \(X\), one can estimate the average treatment effect (ATE) or the average treatment effect on the treated (ATT) as if the treatment assignment were random, mimicking a randomized controlled trial. The positivity assumption, also known as the overlap or support condition, states that every unit (e.g., individual in a study) has a non-zero probability of receiving each treatment level, given the covariates. The consistency assumption states that the potential outcome of an individual under a specific treatment is equal to the observed outcome if the individual receives that treatment. For marketing campaign application, we provide data processing guidelines on how to achieve these assumptions to ensure the validity of causality analysis.

\begin{table}[h!]
\centering
\caption{Causal Data Assumptions: How to satisfy these requirements through data processing.}
\label{table:assumptions_howto}
\begin{tabular}{@{}lp{8cm}@{}}
\toprule
\textbf{Assumption} & \textbf{How to achieve through data processing} \\ \midrule
SUTVA &  Ensure the campaign measurement timeline is short enough before noticeable interferences among units are realized.\\
Ignorability & Control for covariates to achieve independent treatment assignment. \\
Positivity &  Trimming, weighting, or synthetic control group generation to ensure the population data restricts the causality analysis to the overlap region where control and treatment can both be observed. \\
Consistency & Ensure the consistency of treatment or note if there are different versions of treatment. For new campaigns, a new causality analysis needs to be performed. This is important for assessing multiple treatment effects. \\ \bottomrule
\end{tabular}
\end{table}

Once data is pre-processed to avoid yielding biased estimates, we can then apply Algorithm~\ref{alg:causal_segmentation}. The sample result is provided in Section~\ref{s:nr}.

\subsection{Numerical Results}
\label{s:nr}
Before we study the numerical results produced by Algorithm~\ref{alg:causal_segmentation}, we need to ensure the convergence of the algorithm. Below is a sample of the convergence result in Table~\ref{t:convergence}.

\begin{table}[htbp!]
\centering
\caption{Converged Result}
\begin{tabular}{@{}ccccc@{}}
\toprule
ATE & SE & SE-ATE Ratio (\%) & Movement Precision & Segment Movement \\
\midrule
0.507 & 0.005 & 1.053 & 1264.063 & 1235 \\
\bottomrule
\label{t:convergence}
\end{tabular}
\end{table}

In Table~\ref{t:convergence}, the convergence of the causality module produces metrics on ATE, SE, P-Value, SE-ATE Ratio (\%), and Movement Precision. The convergence of the overall movement produces the "Segment Movement" that is lower than "Movement Precision" for the overall system to be considered converged. Additionally, we can visualize the converged segmentation results and sensitivity study of the converged results.

\begin{figure}[ht!]
    \centering
    \begin{subfigure}[b]{0.47\textwidth}
        \includegraphics[width=\textwidth]{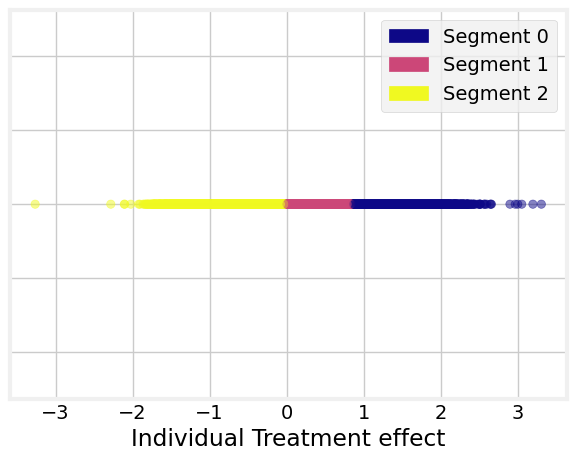}
        \caption{Segmentation Result}
        \label{fig:segresult}
    \end{subfigure}
    ~ 
    \begin{subfigure}[b]{0.51\textwidth}
        \includegraphics[width=\textwidth]{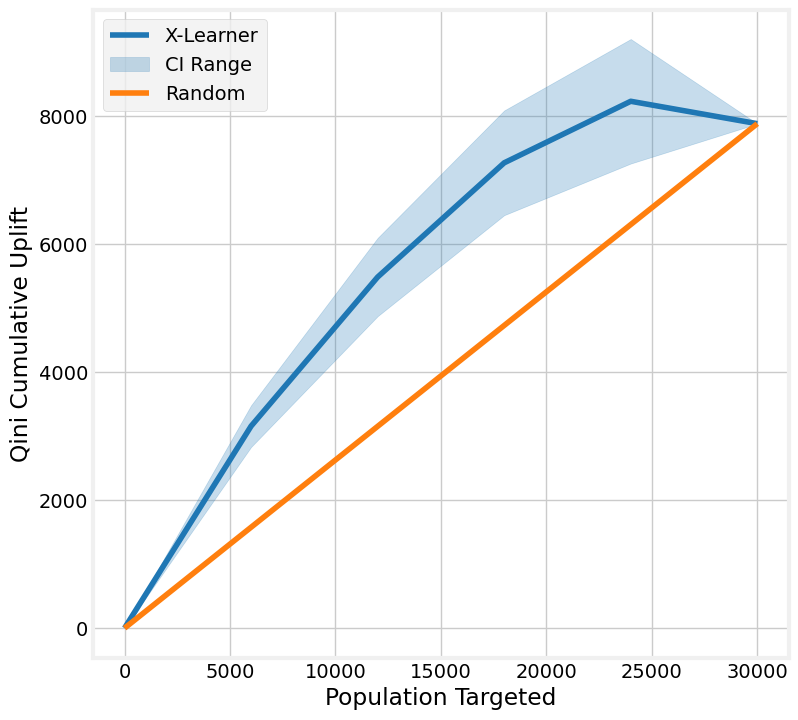}
        \caption{Sensitivity Analysis}
        \label{fig:sensitivity}
    \end{subfigure}
    \caption{Segmentation results and sensitivity study result for the converged system described in Table~\ref{t:convergence}}
    \label{fig:analysisresults}
\end{figure}

Figure~\ref{fig:segresult} shows that the causality is segmented into three segments. For the sensitivity study, we apply the Qini curve measurement~\cite{Radcliffe2011Uplift}. The Qini curve is a performance measurement for uplift modeling, which evaluates the effectiveness of a treatment in a causal inference context. The concept of the Qini curve is analogous to the Gini coefficient used in economics and the ROC curve used in binary classification models, but it is specifically designed for quantifying the incremental impact of a treatment or intervention. Figure~\ref{fig:sensitivity}, measured with a Qini curve, shows that the 90\% confidence range (CI) is shaded. Overall, it appears that even the bottom envelope of our sensitivity study still shows positive improvement over a random assignment Qini curve.

\subsubsection{Simulation Studies and Discussions on KMeans, Propensity Score, and Causal Effect Based Promotion Selection}
\label{s:simulation}
\begin{figure}[h!]
\centering
\includegraphics[width=\linewidth]{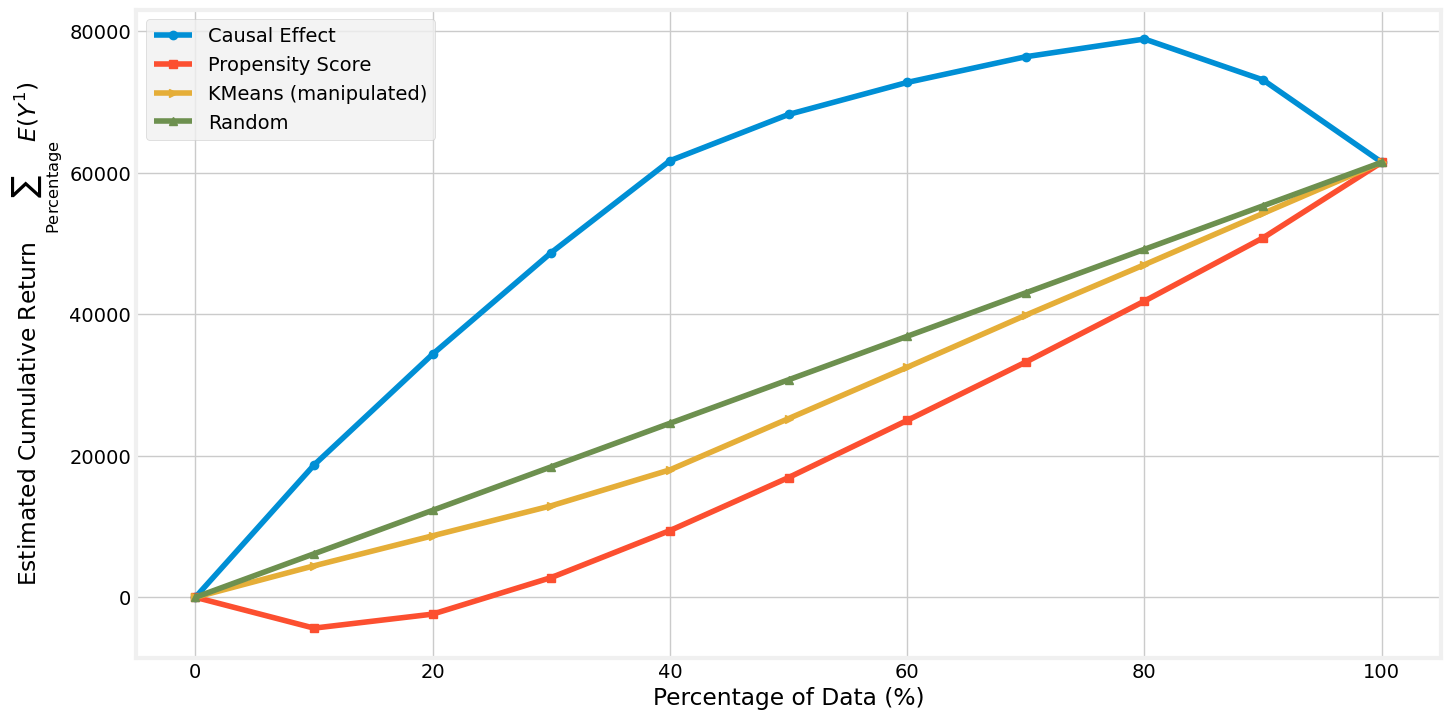}
\caption{Simulation study comparing the performance among the cumulative return gain of four different promotion population selection strategies. Causal Effect selection is based on the treatment effect. Propensity score selection is based on the propensity, which is regressed from the given \(X\) to the propensity of obtaining outcome \(Y\). KMeans is based on clustering results using \(X\). Random selection is a random promotion assignment, whose expected slope is the same as the expected gain, i.e., ATE.}
\label{fig:simulation}
\end{figure}

After causal segmentation converges, we can perform a simulation study following the convergence of Algorithm~\ref{alg:causal_segmentation}. This simulation study compares causality-based population selection, propensity score-based selection, KMeans-based selection, and a random selection strategy. The relative performance of these four selection strategies is graphed in Figure~\ref{fig:simulation}.

Figure~\ref{fig:simulation} is plotted with the population selection percentage on the horizontal axis and the respective cumulative uplift gain on the y-axis. Thus, when \(0\%\) of the population is selected, the overall expected promotion gain is \(0\); when \(100\%\), the selection strategy of any kind no longer matters. Interestingly, the figure shows that the population selected with the causal effect criterion demonstrates the highest overall gain regardless of what percentage of the population is selected for promotions. The KMeans and Random selection curves nearly coincide with each other. Most notably, the promotion target population selected using the propensity score performs even worse than random selection.

\paragraph{Causal Effect}

\begin{figure}[h!]
\centering
\includegraphics[width=\linewidth]{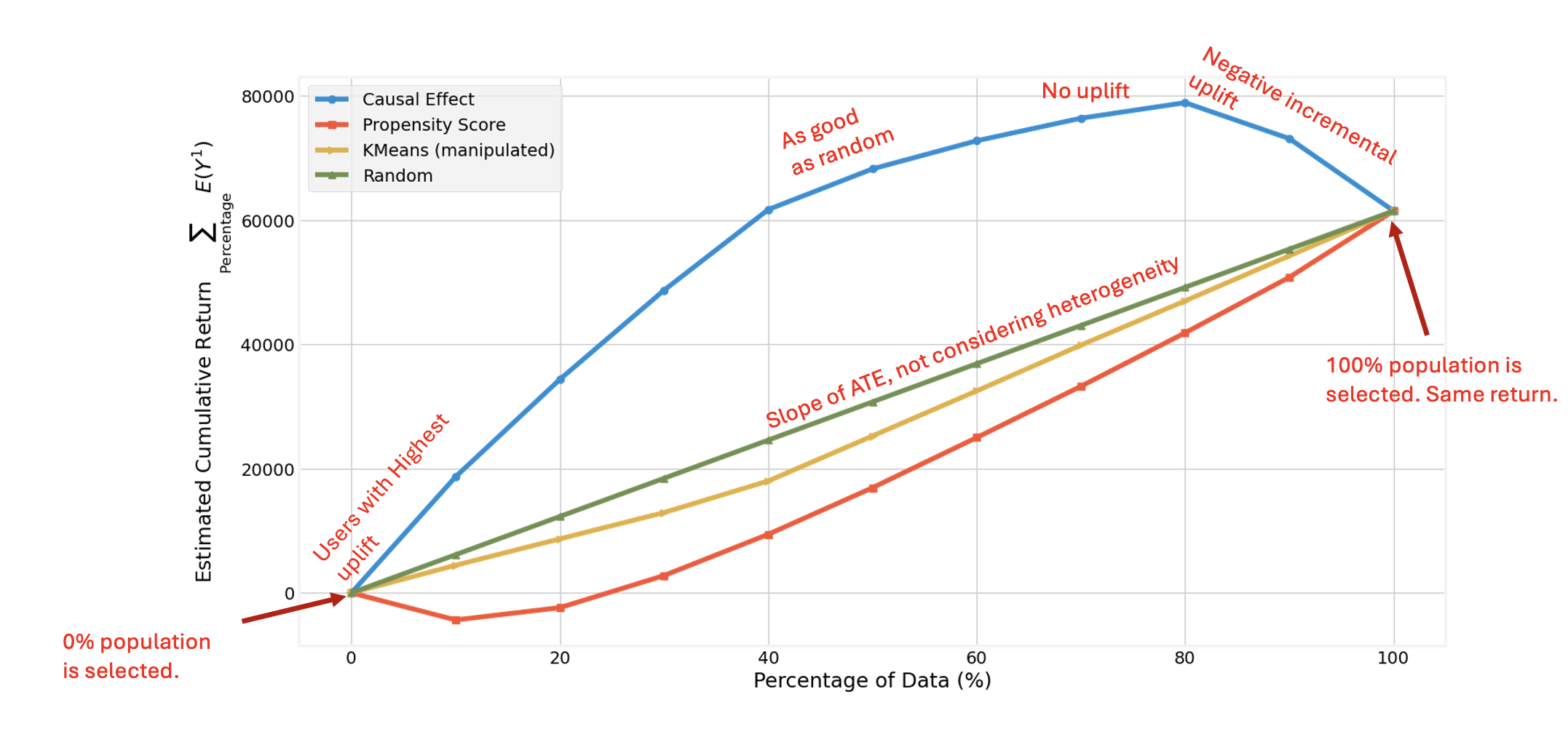}
\caption{Simulation study comparing the performance in terms of cumulative return gain among four different promotion population selection strategies. Specifically, the causal effect curve is labeled.}
\label{fig:simulation_label}
\end{figure}

For more detail on the causal effect selection strategy, this strategy prioritizes individuals based on their ranked uplift effects. We can intuitively understand the convex curve of the causal effect through Figure~\ref{fig:simulation_label}.

Intuitively, selecting a promotion population based on "Causal Effect" (CATE) in descending order is akin to choosing users with the highest uplift first, represented in the lower right corner of Figure~\ref{fig:simulation_label}. The initial slope of the "Causal Effect" is the steepest because it involves selecting users from a small portion of the population. The "Causal Effect" curve will gradually level to match the slope of the "Random" curve, indicating a transition towards random population selection. Eventually, the "Causal Effect" curve may adopt a negative slope, signaling that users with negative uplift are being selected.

\paragraph{K-Means} Why Does Pure K-Means Without Causality Segmentation Perform Poorly? As depicted in Figures~\ref{fig:simulation} and~\ref{fig:simulation_label}, the performance of K-Means aligns closely with that of random assignment strategies for promotion allocation. K-Means is an unsupervised learning algorithm that focuses on partitioning datasets into \(k\) groups based on feature similarity. It aims to divide the \(n\) observations into \(k\) clusters, where each observation is assigned to the cluster with the nearest mean, serving as the prototype for that cluster.

The clusters formed by K-Means are based on the mathematical criterion of minimizing the variance within clusters, as measured by the Euclidean distance. This objective does not necessarily align with human intuition or domain-specific, meaningful groupings. Therefore, while not "artificial" in the sense of being random or arbitrary, the clusters may not always correspond to explainable or expected patterns in the data. K-Means does not ensure that the results of clustering will be inherently understandable or match known categories within the data. The algorithm identifies structures based on its mathematical objective, which may or may not coincide with meaningful or recognizable categories to humans.

K-Means is a powerful tool for exploratory data analysis and can uncover intriguing patterns within the data. However, its simplicity and the nature of its objective function mean that it is most effective under specific conditions—namely, when supplemented by domain knowledge, additional context, or other clustering methods for interpreting the results.

\begin{figure}[ht!]
\centering
\includegraphics[width=\linewidth]{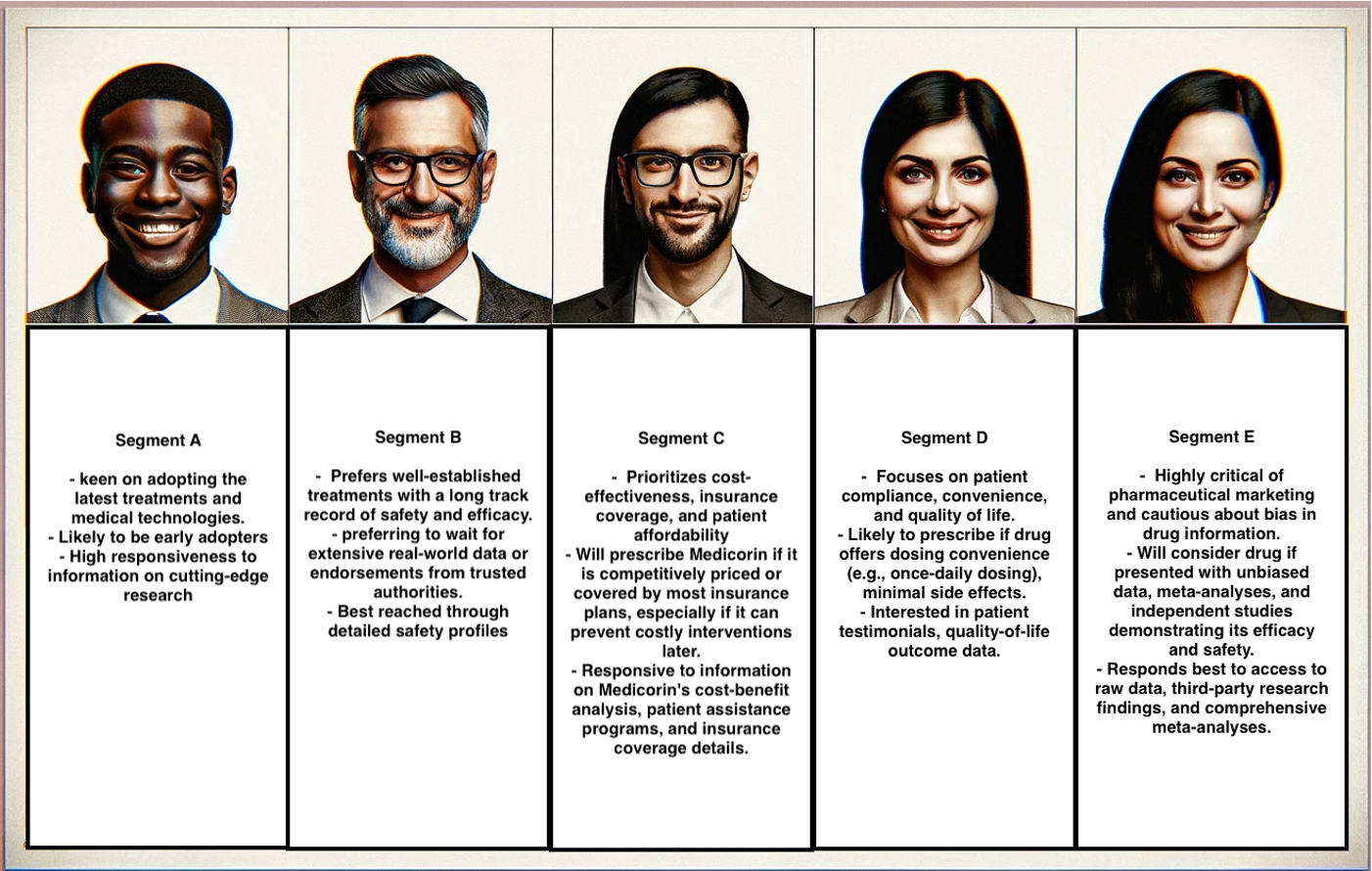}
\caption{When K-Means segmentation is performed for marketing purposes, the segmentation initially relies on input covariates that represent behavior traits. After segmentation is completed, each segment is then characterized by distinct behaviors.\protect\footnotemark}
\label{fig:kmeans}
\end{figure}

Figure~\ref{fig:kmeans} showcases a sample of behavior segmentation results commonly utilized by marketing strategists. The characterization of each K-Means segmented segment is based on a posteriori interpretation rather than predictive outcomes. For instance, Segment A in Figure~\ref{fig:kmeans} is identified as representing loyal, highly interested customers, while Segment E is categorized as comprising highly critical and cautious customers. This segmentation information is relevant for businesses. However, these interpretations are not directly derived from K-Means; the computation steps of K-Means were not informed by these specific objectives or personas. Thus, the explanation of the segments is somewhat artificial, as illustrated in Figure~\ref{fig:kmeans}.

\footnotetext{Disclaimer: Medicorin is a fictional drug name used for illustrative purpose only.}

\paragraph{Propensity Score} The most interesting curve in the simulation studies depicted in Figures~\ref{fig:simulation} and~\ref{fig:simulation_label} relates to the propensity score, which actually performed worse than both the causal effect-based and K-Means-based promotion selection strategies. Propensity models, which assess the relationship between covariates \(X\) and the purchase outcome \(Y\), do not inherently aim to answer how to select a population for promotion activities to maximize uplift gain. Essentially, a propensity model only addresses the likelihood of a purchase occurring given \(X\), not how to select individuals for promotional activities to achieve the greatest uplift.

\begin{figure}[h!]
    \centering
    \includegraphics[width=0.7\textwidth]{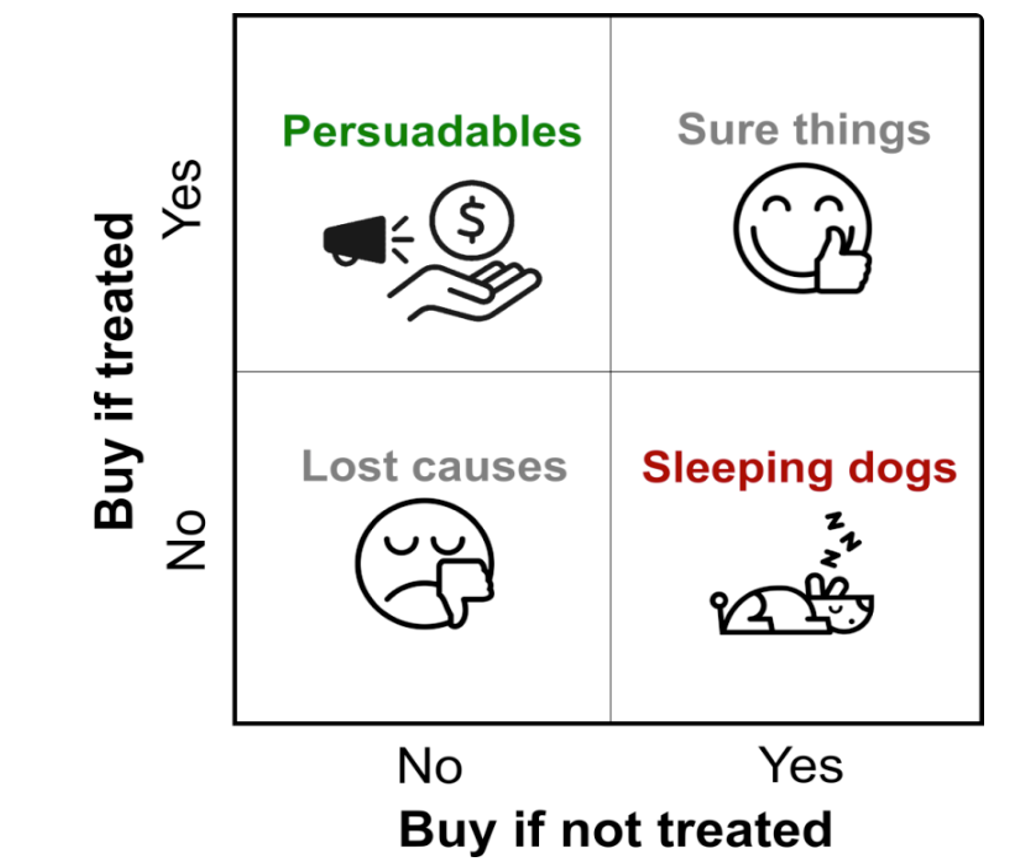}
    \caption{An illustration of quadrants~\cite{Yi2018Pylift}}
    \label{fig:quadrants}
\end{figure}

Figure~\ref{fig:quadrants} illustrates that even if someone has a high propensity to purchase, this should not serve as the sole criterion for deciding whether to allocate marketing resources to specific customers. For example, customers identified as 'Sure things', despite their high propensity, are not ideal candidates for spending marketing dollars on. Conversely, just because customers have a low propensity to purchase does not mean they should automatically be excluded from marketing efforts, especially if they fall into the "Lost Causes" quadrant.

Table~\ref{t:propensity_definition} clearly defines the propensity score as \( P(\text{Purchase } | \text{No Intervention}) \), which is distinct from the objectives addressed by causal effect analysis in Equations~\ref{eqn:ate_math} and~\ref{eqn:cate}. 
\begin{table}[h!]
\centering
\caption{Machine Learning Model Comparison}
\label{t:propensity_definition}
\begin{tabular}{@{}lll@{}}
\toprule
\textbf{ML model} & \textbf{Model tries to answer} & \textbf{Business problem we hope to solve} \\ \midrule
Propensity model & \( P(\text{Purchase } | \text{No Intervention}) \) & Find audiences with: \\
& & High \( P(\text{Purchase } | \text{Intervention}) \) \\
& & Low \( P(\text{Purchase } | \text{Intervention}) \) \\ \addlinespace

Churn model & \( P(\text{Churn } | \text{No Intervention}) \) & Find audiences with: \\
& & High \( P(\text{Churn } | \text{Intervention}) \) \\
& & Low \( P(\text{Churn } | \text{Intervention}) \) \\ \addlinespace

Response model & \( P(\text{Purchase } | \text{Intervention}) \) & Find audiences with: \\
& & High \( P(\text{Purchase } | \text{Intervention}) \) \\
& & Low \( P(\text{Purchase } | \text{Intervention}) \) \\ \bottomrule
\end{tabular}
\end{table}

Therefore, neither the churn model \(P(\text{Churn } | \text{No Intervention})\) nor the response model \(P(\text{Purchase } | \text{Intervention})\) should be used as strategies for selecting customers for promotions.

\begin{figure}[h!]
\centering
\includegraphics[width=.8\linewidth]{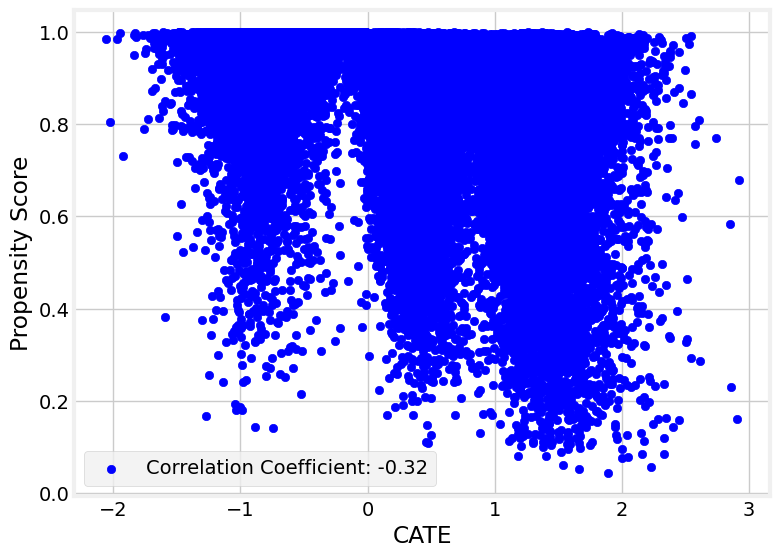}
\caption{Not only are they not correlated, they are mildly negatively correlated.}
\label{fig:correlation}
\end{figure}

However, this does not fully explain why the propensity score selection strategy is the least effective among all promotional target selection strategies depicted in Figures~\ref{fig:simulation} and~\ref{fig:simulation_label}. To clarify, we examine the relationship between propensity scores and CATE, as illustrated in Figure~\ref{fig:correlation}.

Figure~\ref{fig:correlation} shows that propensity scores and CATE are not strongly correlated; in fact, they are weakly negatively correlated. Therefore, propensity scores cannot serve as a substitute for causality analysis in uplift modeling.

\subsection{Explainability}

Given the exclusivity between causality analysis and segmentation results as outlined in Theorem~\ref{thmd:exclusivity}, the explainability of the overall iterative causal segmentation algorithm merits discussion. Although the segmentation algorithm is unsupervised and ordered by thresholding, Algorithm~\ref{alg:causal_segmentation} can be elucidated through its causality module. This module records the converged state when all three modules—causality, segmentation, and segment movement—have converged. Techniques like SHAP values can be adopted to provide granular explanations.

\begin{figure}[h!]
\centering
\includegraphics[width=.8\linewidth]{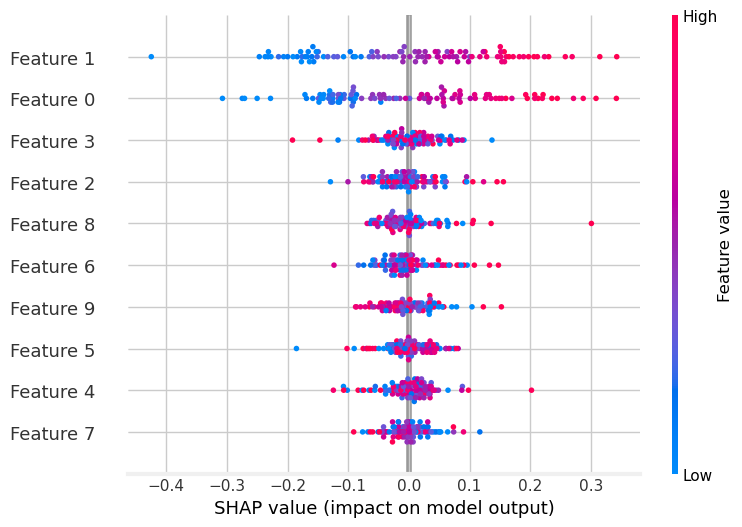}
\caption{Example of SHAP value explainability for the converged results of Table~\ref{t:convergence} with a subpopulation of 100 data points. Note: SHAP value computation can be computationally expensive.}
\label{fig:shap}
\end{figure}

Since SHAP values can offer individualized explanations regarding features, as demonstrated in Figure~\ref{fig:shap}, they can be utilized for the causality module of the iterative causal segmentation algorithm (Algorithm~\ref{alg:causal_segmentation}). Moreover, due to the exclusivity detailed in Theorem~\ref{thmd:exclusivity}, the SHAP value explainability for the causality module also extends to the overall explainability of the iterative causal segmentation algorithm.

\section{Conclusion}

In this paper, we have introduced the iterative causal segmentation algorithm, Algorithm~\ref{alg:causal_segmentation}, designed specifically for marketing contexts where segmentation strategies play a crucial role in influencing purchase outcomes. This necessitates addressing the tightly coupled system between promotion and segmentation.

We demonstrated the value of this segmentation algorithm by comparing it with other common machine learning models used in marketing settings in Section~\ref{s:simulation}. Furthermore, we established the exclusivity of the proposed segmentation method, showing that it cannot be easily replaced by other methods, as evidenced in Section~\ref{pf:exclusitivity}.

\bibliographystyle{plain}
\bibliography{references} 

\end{document}